\documentclass[smallabstract,smallcaptions]{dccpaper}

\usepackage{epsfig}
\usepackage{amsmath}
\usepackage{amssymb}
\usepackage{color}
\usepackage{url}

\newlength{\figurewidth}
\newlength{\smallfigurewidth}

\usepackage{subcaption}
\usepackage{makecell}
\usepackage{multirow}
\usepackage{array}
\usepackage{booktabs}
\usepackage{tabularx}
\usepackage{amsmath}
\usepackage{bm}
\usepackage{mathrsfs}
\usepackage{hyperref}
\begin{document}

\title
{\large
\textbf{Hierarchical Attention Networks for Lossless Point Cloud Attribute Compression}
}

\author{%
Yueru Chen$^{\dag}$, Wei Zhang$^{\dag}$, Dingquan Li$^{\dag}$, Jing Wang$^{\dag}$, and Ge Li$^{\ast}$\\[0.5em]
{\small\begin{minipage}{\linewidth}\begin{center}
\begin{tabular}{ccc}
$^{\dag}$Pengcheng Laboratory, Shenzhen, China \\
$^{\ast}$Peking University Shenzhen Graduate School, Shenzhen, China 
\end{tabular}
\end{center}\end{minipage}}
}

\maketitle
\thispagestyle{empty}

\begin{abstract}
In this paper, we propose a deep hierarchical attention context model for lossless attribute compression of point clouds, leveraging a multi-resolution spatial structure and residual learning. A simple and effective Level of Detail (LoD) structure is introduced to yield a coarse-to-fine representation. To enhance efficiency, points within the same refinement level are encoded in parallel, sharing a common context point group. By hierarchically aggregating information from neighboring points, our attention model learns contextual dependencies across varying scales and densities, enabling comprehensive feature extraction. We also adopt normalization for position coordinates and attributes to achieve scale-invariant compression. Additionally, we segment the point cloud into multiple slices to facilitate parallel processing, further optimizing time complexity. Experimental results demonstrate that the proposed method offers better coding performance than the latest G-PCC for color and reflectance attributes while maintaining more efficient encoding and decoding runtimes.
\end{abstract}

\section{Introduction}
The increasing commercial demand for 3D point clouds has sparked growing interest in academia and industry.
A point cloud consists of a large number of points in 3D coordinates, each associated with corresponding attributes (e.g., colors, reflectances), resulting in substantial storage requirements and high transmission costs. Therefore, efficient point cloud compression has become an essential and valuable research topic. 
Compared to traditional image and video data, point clouds exhibit highly irregular structures with varying densities and scales, making it challenging to effectively capture correlations between points. Moreover, the diversity of attribute types adds further difficulties in developing generic attribute compression schemes. This paper aims to provide a comprehensive solution for lossless point cloud attribute compression (PCAC).

\subsection{Background and Challenges}
Conventional hand-crafted point cloud attribute compression has been extensively studied. 
One effective solution is based on the Graph Fourier Transform (GFT), which models local connectivity and spatial relationships using graphs, allowing for the decorrelation of attributes via graph transform matrices~\cite{zhang2014point,zhang2024efficient}.
Considering both coding performance and computational efficiency, the state-of-the-art scheme for PCAC is in the MPEG G-PCC~\cite{graziosi2020overview} reference software TMC13, which includes two attribute coding tools: Pred-Lifting (PLT) and predictive RAHT (PRAHT). 
PLT~\cite{Mammou2018Lifting} is an enhanced lifting framework built on the Level of Detail (LoD) structure, while PRAHT integrates predictive techniques into the region adaptive hierarchical transform (RAHT)~\cite{de2016compression}.
However, the hand-crafted approach has several drawbacks: it relies on manually designed graphs and transform matrices, which may not fully capture complex and diverse geometric patterns; additionally, it makes a strong assumption of a high correlation between geometry and attributes, a condition that may not always hold in real-world scenarios.

Numerous studies have explored the application of deep learning methods in lossless PCAC~\cite{quach2022survey}. 
Wang~\textit{et al.}~\cite{3cacwang2023lossless} developed an end-to-end point cloud attribute compression framework extending the multiscale structure from SparsePCGC. It uses a sparse CNN to estimate Laplacian distribution parameters for attribute probability derivation.
Nguyen~\textit{et al.}~\cite{nguyen2023lossless} built an autoregressive context model to directly learn the probability density function of attributes through a sparse tensor-based neural network.
However, existing learning-based lossless PCAC methods have several limitations. One key issue is their poor generalization to point clouds with varying scales and densities. Moreover, most of these methods rely on autoregressive context models, resulting in high time complexity. 

In contrast to traditional image and video formats organized in regular dense grids, point clouds are irregularly distributed in 3D space. For efficient processing, both hand-crafted and learning-based methods typically employ hierarchical structures to organize the points. The octree is a simple and effective structure for point clouds and has been widely adopted~\cite{chen2020point,fu2022octattention}. The MPEG G-PCC utilizes another geometry structure known as Level of Detail (LoD)~\cite{schwarz2018emerging}, which reorganizes the input point cloud into a set of refinement levels. However, the LoD used in G-PCC relies on Euclidean distance calculations, leading to higher computational complexity and a point-wise autoregressive coding process that limits the efficiency of parallel computing.

\subsection{Our Approach and Contributions}
This paper introduces a hierarchical attention network for lossless point cloud attribute compression, referred to as HA-PCAC, to address the aforementioned challenges. We first develop a simple and fast Level of Detail (LoD) construction method. Unlike G-PCC, we utilize permutation distances based on the Hilbert index, consequently reducing computational complexity while generating coarse-to-fine refinement levels of the point cloud. Additionally, we allow points within the same refinement level to be encoded simultaneously by sharing a common context point group. 
Next, a deep hierarchical attention context model is proposed to estimate the probability distribution of attributes. The hierarchical structure is designed to learn from broader neighborhoods of points in unstructured space and effectively capture contextual relationships across different spatial scales. In contrast to most learning-based methods that directly learn from attribute values, the proposed HA-PCAC utilizes residual attributes as input, facilitating better learning of finer details and faster convergence during the training process. A simple interpolation-based attribute predictor is used to calculate these residuals.

The main contributions of this paper are summarized as follows: 1) We propose a hierarchical attention network that learns broader spatial contexts and extracts effective features through a level-wise autoregressive coding process, resulting in more effective lossless attribute compression performance; 2) By leveraging the normalization of positions and attributes, the proposed HA-PCAC generalizes well across point clouds with varying densities, geometric scales, and attribute scales; 3) The experimental results demonstrate the state-of-the-art performance on various large-scale datasets for both color and reflectance attributes.

\section{Method}
\begin{figure*}[!t]
\centerline{\includegraphics[width=0.95\linewidth]{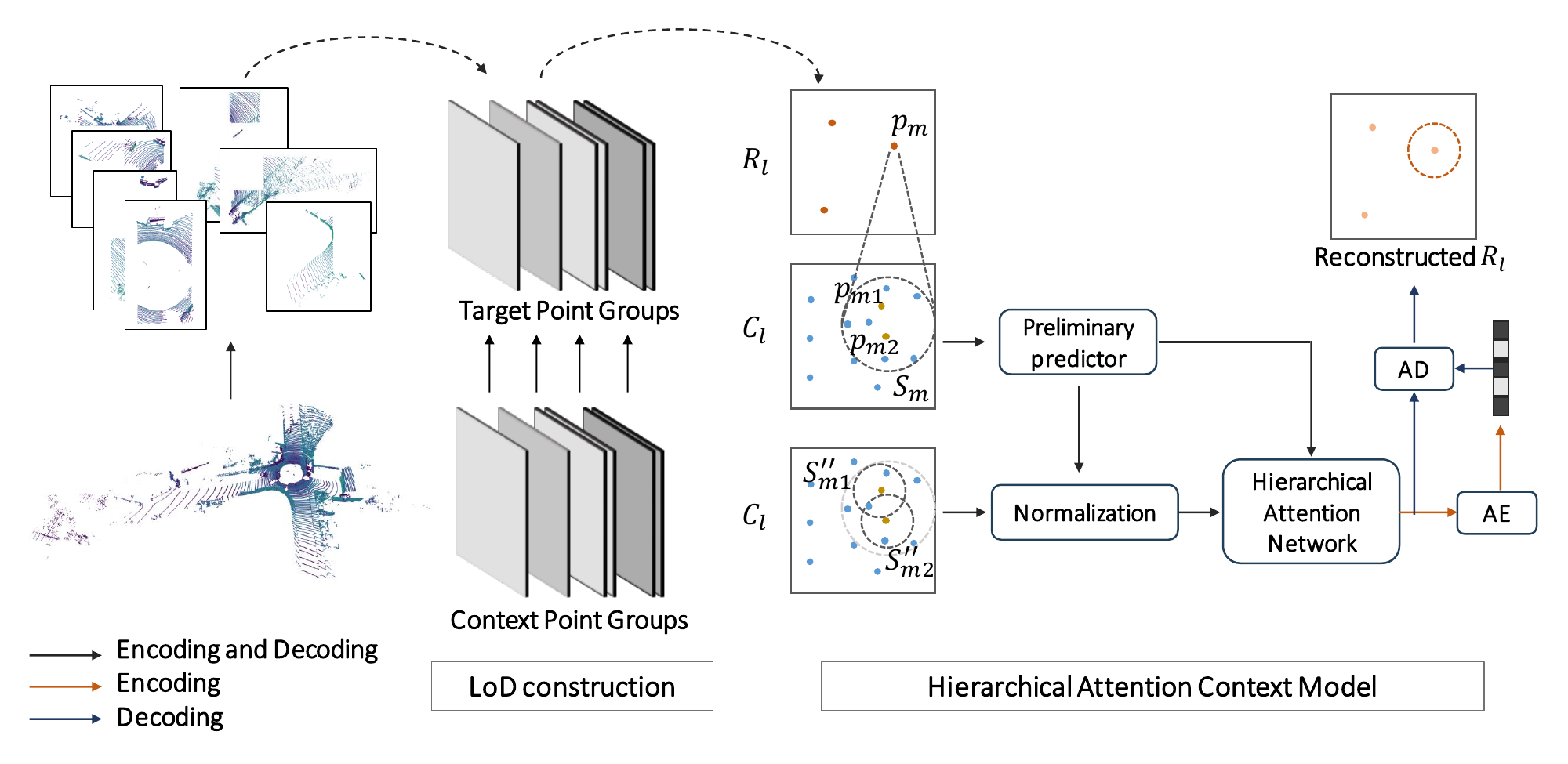}}
\caption{The point cloud is first partitioned into multiple slices. The points in each slice are then organized into several target groups with resolutions ranging from coarse to fine using our fast Level of Detail (LoD) construction. A deep hierarchical attention context model is employed to estimate the probability distribution of the attributes.}\label{fig:overview}
\end{figure*}

The proposed hierarchical attention network for lossless point cloud attribute compression (HA-PCAC) is introduced in this section. We assume that the point cloud geometry information is available in advance, and the points are organized in Hilbert order~\cite{lawder2001querying}.
\subsection{Framework Overview}
The framework overview of the proposed HA-PCAC is illustrated in Fig.~\ref{fig:overview}. 
We first utilize a point cloud slicing technique that partitions the points according to their Hilbert order with a predetermined number per slice; each slice is then compressed independently. 
To further enhance parallelism, the HA-PCAC processes points at the same refinement level simultaneously, leveraging the designed LoD structure.
In addition, we introduce a deep hierarchical attention network that extracts sufficient contextual features with larger receptive fields for improved compression performance and incorporates residual learning to accelerate the convergence of network training.
 
\subsection{Fast Level of Detail (LoD) Generation}
This subsection introduces a fast LoD construction method that enables progressive refinement with minimal computational overhead. The point cloud is reorganized into a set of refinement levels based on permutation distances derived from the Hilbert index, and the corresponding context point groups are gathered.
\begin{figure}[!t]
    \centering
    \begin{minipage}[t]{0.61\textwidth}
        \centering
        \includegraphics[width=0.95\textwidth]{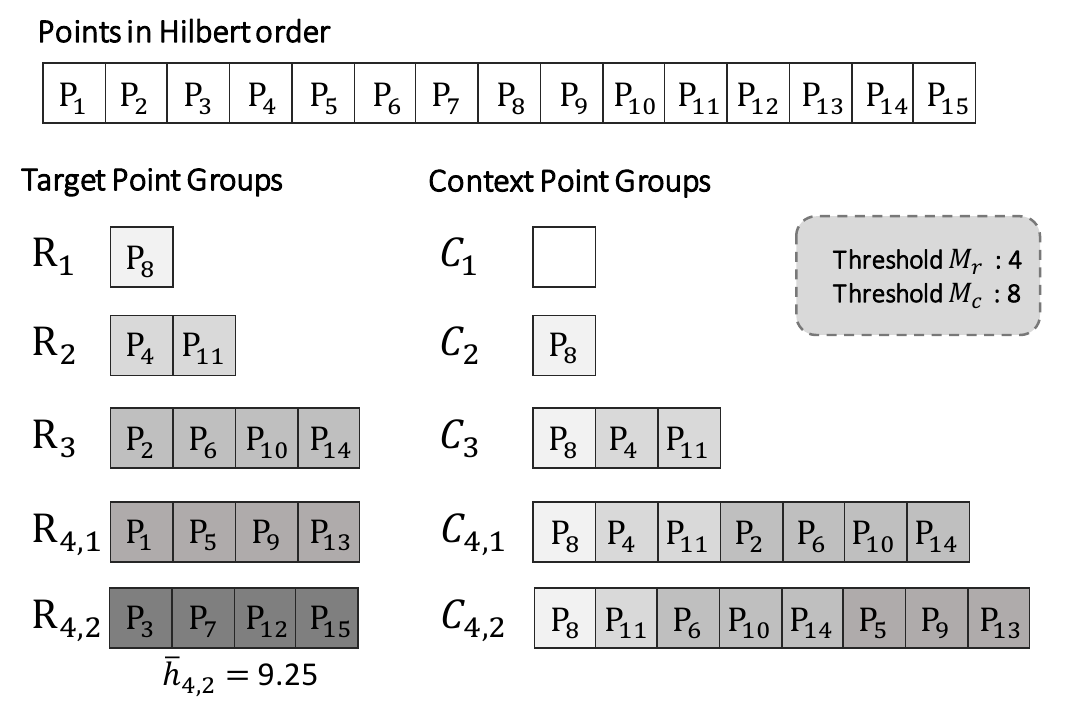}
        \caption{Illustration of fast LoD construction.}
        \label{fig:lod}
    \end{minipage}%
    \hfill
    \begin{minipage}[t]{0.37\textwidth}
        \centering
        \includegraphics[width=0.9\textwidth]{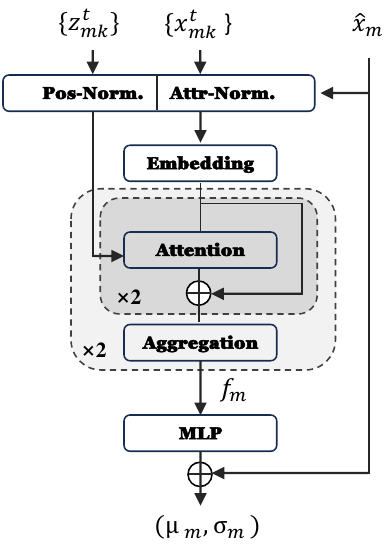}
        \caption{Overall Architecture.}
        \label{fig:atten}
    \end{minipage}
\end{figure}
We define a point cloud of $N$ points as ${\bm P}=\{{\bm p}_1,\dots,{\bm p}_N\}$. Assume the point cloud ${\bm P}$ contains coordinates ${\bm Z} = \{{\bm z}_i\}$, attributes ${\bm X} = \{{\bm x}_i\}$, and corresponding Hilbert indices ${\bm H} = \{h_i\}$. The number of points at each LoD level is predefined as \( \{n_1, \dots, n_L\} \), with \( n_1 = 16 \) and \( n_l = 2 \times n_{l-1} \) for the experiments. To manage memory usage effectively, we limit the maximum number of points per level by setting a threshold \( M_r \). For $l = 1,\dots,L$, the point collections at refinement level $l$ are $\{{\bm R}_l\}$, acquired as follows:

A non-selected point list, denoted as ${\bm P_{ns}}$, is initially set to ${\bm P_{ns}} = {\bm P}$.
During the construction process, points in ${\bm P_{ns}}$ are arranged according to their Hilbert indices, and the construction is completed when ${\bm P_{ns}}$ becomes empty.

\begin{itemize}
    \item \textbf{Constructing \( {\bm R_l} \).}
    First, calculate the selection interval length $K_l$ as $K_l = \frac{|{\bm P_{ns}}|}{n_l}$. The first point of \( {\bm R_l} \) is randomly selected from the first $K_l$ points in ${\bm P_{ns}}$. The remaining \( n_l - 1 \) points of \( {\bm R_l} \) are then chosen at equal intervals from ${\bm P_{ns}}$. Finally, update ${\bm P_{ns}} = {\bm P_{ns}} - {\bm R_l}$.
    \item \textbf{Constructing \( {\bm R_L} \).}
    If the number of points in ${\bm P_{ns}}$ is less than the preset \( n_L \), we set ${\bm R_L} = {\bm P_{ns}}$ and update ${\bm P_{ns}} = \emptyset$. 
    \item \textbf{Splitting \( {\bm R_l} \).}
    If \( n_l > M_r \), \({\bm R_l}\) is further divided into multiple sub-levels \({\bm R_{l,s}}\), each containing \( M_r \) points.  It should be noted that the points within \({\bm R_l}\) are sorted, and each sub-level \({\bm R_{l,s}}\) is compressed in the same manner as an entire refinement level.
\end{itemize}

Through this approach, we achieve rapid LoD construction acquiring a coarse-to-fine refinement level \( {\bm R_1} \) to \( {\bm R_L} \). Correspondingly, we pre-construct level-wise context point groups, denoted as \( \{{\bm C_1},\dots, {\bm C_L}\}\). 
To control memory usage and computational load, we limit the size of the context point group ${\bm C_l}$ by setting a maximum threshold of \( M_c \).
Start with \({\bm C_1} \) as an empty set and initialize \( {\bm C_l} = \sum_{j=1}^{l-1} {\bm R_j} \). If the number of points in \( {\bm C_l} \) exceeds \( M_c \), we update \( {\bm C_l} \) by selecting the \( M_c \) points that are closest to \( {\bm R_l} \). As illustrated in Fig.~\ref{fig:lod}, the average Hilbert index \( {\overline{h}_l} \) of \( {\bm R_l} \) is computed, and based on the index distances, the nearest \( M_c \) points are chosen. 

Compared to conventional LoD structures, our approach simplifies calculations by using a simple order distance metric. It also offers the advantage of coarse-to-fine point cloud representations, ensuring that the constructed context sets contain a more balanced and comprehensive spatial range of reference points. 

\subsection{Hierarchical Attention Networks}

\subsubsection{Problem Formulation}

To solve lossless PCAC, we estimate the ground truth distribution $P({\bm X})$ using a probability distribution $P_\theta({\bm X})$ derived from the proposed hierarchical attention context model with a level-wise autoregressive coding process. Compared to sequential point-by-point encoding approaches, the proposed HA-PCAC enables the parallel processing of points within the same refinement level by leveraging a shared context set.
Mathematically, the model is expressed as:
\begin{align}
    P_\theta({\bm X}) &= \prod\limits_{i} p_\theta({\bm x_i}|{\bm x_1},{\bm x_2},\ldots,{\bm x_{i-1}};{\bm z_1},{\bm z_1},\ldots,{\bm z_{i}}) \nonumber \\
    &= \prod\limits_{l}\prod\limits_{m} p_\theta({\bm x_{lm}}|{\bm C_l};{\bm z_{lm}})  \nonumber \\
    &= \prod\limits_{l} p_\theta({\bm R_{l}}|{\bm C_l};{\bm R_{l}^z})    
\end{align}
where ${\bm C_l}$ and ${\bm R_{l}^z}$ denote the context point group and the coordinates of the target point group at refinement level $l$, respectively; ${\bm x_{lm}}$ and ${\bm z_{lm}}$ denote the attributes and positions of the $m$th point in the $l$th target group, respectively.
We directly save the attributes of the first target group ${\bm R_1}$. When $l>1$, we model the probability distribution of attributes using a Laplace distribution $\mathscr{L}$, with the attribute $\bm x_{i}$ estimated by parameters ${\bm \mu_i}$ and ${\bm \sigma_i}$, which are derived from a network \( g_\theta(\cdot) \). The estimation for the $l$th level is expressed as:
\begin{align}
    \prod\limits_{m} p_\theta({\bm x_{lm}}|{\bm C_l};{\bm z_{lm}}) &= \prod\limits_{m} \int_{{\bm x_{lm}}-\frac{1}{2}}^{{\bm x_{lm}}+\frac{1}{2}}
    \mathscr{L}(y|{\bm \mu_{lm}},{\bm \sigma_{lm}})\mathrm{d}y  \nonumber \\
    &= \prod\limits_{m} \int_{{\bm x_{lm}}-\frac{1}{2}}^{{\bm x_{lm}}+\frac{1}{2}}
    \mathscr{L}(y|g_\theta({\bm C_l};{\bm z_{lm}}))\mathrm{d}y. 
\end{align}

Finally, the loss of the training model is measured in terms of total bits:
\begin{equation}
Loss = \sum_{i=1}^{N} bits({\bm x_{i}}) = -\sum_{i=1}^{N} \log_2(p_\theta(\bm x_i))=-\sum_{l=1}^{L}\sum_{m} \log_2(p_\theta(\bm x_{lm})).
\end{equation}

\subsubsection{Hierarchical Attention Context Model}
This section details the proposed context model \( g_\theta \). As shown in Fig.~\ref{fig:atten}, we design a two-stage hierarchical attention network that integrates residual learning to effectively capture broader spatial patterns and finer details.
The features extracted by attention modules are subsequently fed into a multilayer perceptron (MLP) to estimate the location and scale parameters of the Laplacian distribution.

As an example, we present the compression process of the target point group \( \bm R_l \) conditioned on the context point group \( \bm C_l \). For simplicity, we omit the subscripts $l$ of the variables, which represent the index of the refinement level. 
For each point \( \bm p_m \) in \( \bm R_l \), we use the k-nearest neighbors (KNN) algorithm to identify the closest \( K \) points from \( \bm C_l \), which are sorted by Euclidean distance from nearest to farthest, forming the context point subgroup \( \bm S_m \). 
We utilize an interpolation-based method as the preliminary predictor to estimate the attribute value of \( p_m \), denoted as \( \bm{\hat{x}_m} \), based on inverse distance weighting. The first three neighbor points in \( \bm S_m \) are selected as reference neighbors. 
\begin{equation}
\bm{\hat{x}_m} = \frac{\sum_{\bm p_j \in S_m} w_{mj} \cdot \bm x_j}{\sum_{\bm p_j \in S_m} w_{mj}}, \quad \quad w_{mj} = \frac{1}{\| {\bm z}_j - {\bm z}_m \|}.
\end{equation} 

Next, we designate the first $K_1$ points in ${\bm S_m}$ as the neighbor point group of \( {\bm p}_m \), referred to as ${\bm S'_m}=\{{\bm p}_{mk}\}$, where $k=1,\dots,K_1$. 
For each \( \bm{p}_{mk} \), we then find its nearest \( K_2 \) points within the context subgroup \( \bm{S}_m \), forming the point collections \( \bm{S}''_{mk} = \{ \bm{p}_{mk}^t \} \), for \( t = 1, \dots, K_2 \). We set the point ${\bm p}_{mk}$ to be the first element ${\bm p}_{mk}^{1}$ in ${\bm S''_{mk}}$. 

Utilizing the network structure illustrated in Fig.~\ref{fig:atten}, the input to the first stage of the attention module is point groups $\{{\bm S''_{mk}}\}$, and the feature vectors ${\bm f_{mk}}$ of the local region around the neighbor point ${\bm p_{mk}}$ is extracted. 
The normalization is applied separately to the coordinates $\{{\bm z_{mk}^{t}}\}$ and attribute values $\{{\bm x_{mk}^{t}}\}$ to achieve scale invariance, formulated as:
\begin{equation}
\bm {\bar{z}_{mk}^{t}} =\frac{\bm {z_{mk}^{t}}-\bm {z_{mk}^{1}}}{\max\limits_{2 \leq q \leq K_2} \left\{ \lVert \bm {z_{mk}^{q}} - \bm {z_{mk}^{1}} \rVert \right\}}, \quad \quad {\bm {\bar{x}_{mk}^{t}}} =\frac{\bm x_{mk}^{t}-\bm {\hat{x}_m}}{MAX_{attri}},
\end{equation} 
where $MAX_{attri}$ are the maximum attribute values. For color attributes, we perform compression calculations in the YCoCg color space, with maximum values of 256 for luminance and 512 for chrominance.
The role of normalization is to ensure that the point positions within the neighborhood of ${\bm S''_{mk}}$ are centered around the neighbor point ${\bm p}_{mk}$, while the attributes are centered around the target point ${\bm p}_{m}$. From another perspective, attribute normalization can be viewed as a form of residual learning. 
Instead of learning from attribute values, our method explicitly models the attribute residuals. Experimental results confirm the faster convergence of the loss as shown in Fig.~\ref{fig:convergence}.

We employ an attention mechanism that incorporates position embedding and subtraction relations, as proposed in PTv2~\cite{wu2022point}, and the attention score can be formulated as follows:

\begin{equation}
{\bm{s}_{mk}^{t}} = s\left(\delta_{mul}(\bm {\bar{z}_{mk}^{t}}) \odot (\psi_{key}(\bm f_{mk}^{t})-\psi_{query}(\bm f_{mk}^{1}))+\delta_{bias}(\bm {\bar{z}_{mk}^{t}})\right),
\label{eq:score}
\end{equation}  
where $s$, $\delta_{mul}$, $\delta_{bias}$, $\psi_{key}$ and $\psi_{query}$ are all  MLPs. $s$ is for weight encoding, $\delta_{mul}$, $\delta_{bias}$ handle position coding, while $\psi_{key}$ and $\psi_{query}$ project point feature to query and key, respectively. The embedding feature vector of point ${\bm p}_{mk}^{t}$ is denoted as $\bm f_{mk}^{t}$.
The final representation ${\bm f_{mk}}$ for neighboring point ${\bm p}_{mk}$ is derived by aggregating the output features from the first attention block using summation.

In the second stage, we take the coordinates $\{{\bm z_{mk}}\}$ and feature vectors $\{{\bm f_{mk}}\}$ of point group ${\bm S'_{m}}$ as input.
Similarly, we normalize the coordinates by centering around the target points ${\bm p}_{m}$ and scaling using the farthest neighbor distance. 
We maintain the assumption that the features from the first stage are already centered around the target point, eliminating the need for additional attribute normalization.
Sequentially, the query should be set to zero in Eq.~\eqref{eq:score}. Finally, features vector $\bm f_{m}$ of the target point is obtained. 

To estimate the location and scale parameter (${\bm \mu_m}$, ${\bm \sigma_m}$) of the Laplace distribution for each point ${\bm p}_{m}$, the features derived from the attention module are fed into an MLP. Due to residual learning, it is necessary to add the predicted values back as follows:
\begin{equation}
(\bm {\mu'_m} , \bm {\sigma_m}) = MLP(\bm f_m), \quad \quad
\bm{\mu_m} = \bm {\mu'_m} + \bm {\hat{x}_m}.
\end{equation} 

\section{EXPERIMENTS}
\subsection{Experimental Setup}

\textbf{Dataset.} We conducted extensive experiments using various datasets to validate the robustness and effectiveness of the proposed HA-PCAC method. 
For color attributes, we test on the 8i Voxelized Full Bodies (8iVFB)~\cite{dEon2017voxelized}, a widely used dataset of human body sequences. We use the ``longdress" and ``soldier" sequences for training, and the ``loot" and ``redandblack" sequences for testing. Additionally, we evaluate ScanNet~\cite{dai2017scannet}, a large-scale 3D indoor scene dataset of real-world environments captured with depth sensors. Following the recommended splits, the training dataset includes 1,503 scans, while the testing dataset includes 100 scans.
For reflectance attributes, we utilize the SemanticKITTI dataset~\cite{geiger2012we}, which contains more than 43,000 scans of LiDAR-acquired point clouds from real-world autonomous driving scenes, featuring elements such as vehicles and buildings. Adhering to the standard train-test splits, sequences 00 to 10 are used for training, and sequence 11 is reserved for testing. The reflectance values are quantized to 8 bits for training.

\noindent\textbf{Baseline.} We conduct performance benchmarking of our method against the latest publicly-available G-PCC test model TMC13v23~\cite{TMC13v23}, comparing both the PLT and PRAHT schemes. 
We evaluate lossless compression performance by reporting the bits per point (bpp) of attributes and the time complexity.

\noindent\textbf{Implementation Detail.}
We train and test our model on the NVIDIA Tesla T4 GPU and Intel Xeon Gold 6248 Processor (27.5M Cache, 2.50 GHz). We set the maximum size thresholds $M_r$ and $M_c$ to 512 and 2048, respectively, while the context point subgroup size $K$ is 32, and the number of neighbor points $K_1$ and $K_2$ are both set to 8. The slicing technique is applied to all datasets, where the point clouds are divided into slices of $2^{14}$ points.

\subsection{Experimental Results}
We present a comparison of lossless compression performance in Table~\ref{table:bpp1}. Compared to the second-best method (PLT), our method achieves average bpp reductions of 4.75\% on the i8VFB dataset. For ScanNet and SemanticKITTI, we utilize a training set with 12-bit coordinates, evaluating both 12-bit and 8-bit testing sets. Table~\ref{table:bpp1} shows consistent performance improvements over baseline methods. When evaluating against PLT, our proposed method achieves gains of 6.23\% and 5.99\% on ScanNet, as well as 12.01\% and 13.99\% on SemanticKITTI, for the 12-bit and 8-bit testing sets, respectively.
The results indicate that our method performs better on sparser datasets and generalizes well across different geometric scales, owing to the hierarchical network architecture effectively capturing features from larger spatial regions.
\begin{table*}[!t]
  \centering
  \caption{Comparisons of coding performance (bpp) and time complexity
  (s/frame)}
  \label{table:bpp1} 
  
  \begin{tabular}{|l|c|c|c|c|}
    \hline
    \multicolumn{2}{|c|}{\multirow{1}{*}{\textbf{Test Data}}} &
    \textbf{PLT} & \textbf{PRAHT} & \textbf{HA-PCAC} \\
    \hline
    \multirow{3}{*}{8iVFB}& {loot} & 6.08 & 7.00 & 5.73  \\
     & {redandblack} & 9.15 & 9.59 & 8.80  \\
    \hline
    \multirow{2}{*}{ScanNet}& 12bit & 13.21 & 13.45 & 12.39 \\
    \cline{2-5}
     & 8bit & 13.67 & 13.88 & 12.85  \\
    \hline
    \multirow{2}{*}{Kitti}& 12bit & 6.62 & 6.63 & 5.83  \\
    \cline{2-5}
     & 8bit & 6.83 & 6.85 & 5.88  \\
    \hline
    \multicolumn{5}{|c|}{\multirow{1}{*}{\textbf{Avg. Times (Encoding / Decoding)}}} \\
    \hline
    \multicolumn{2}{|c|}{8iVFB} & 8.9/8.7 & 3.4/3.4 & 1.5/1.4 \\
    \multicolumn{2}{|c|}{ScanNet\_12bit} & 2.1/2.1 & 1.5/1.4 & 1.4/1.4 \\
    \multicolumn{2}{|c|}{Kitti\_12bit} & 1.0/1.0 & 0.5/0.4 & 0.8/0.8 \\
    \hline
  \end{tabular}
\end{table*}
\begin{figure}[!t]
    \centering
    \begin{minipage}{0.59\textwidth}
        \captionof{table}{Comparisons of the proposed HA-PCAC and G-PCC-PLT with varying downsample step sizes and different reflectance quantization precisions}
        \label{table:ex2}
        \centering
        \begin{tabularx}{0.985\textwidth}{|c|c|c|c|c|}
            \hline
            \multirow{2}{*}{\textbf{Step Sizes}} & \multicolumn{2}{c|}{\textbf{ScanNet\_12bit}} & \multicolumn{2}{c|}{\textbf{Kitti\_12bit}} \\
            \cline{2-5} & bpp & gain & bpp & gain \\
            \hline
            2 & 12.43 & 7.60\% & 6.03 & 11.57\% \\
            8 & 12.54 & 6.10\% & 6.03 & 12.44\% \\
            32 & 13.68 & 6.04\% & 5.68 & 16.64\% \\
            \hline
            \multirow{2}{*}{\textbf{Refl. Bits}} & \multicolumn{4}{c|}{\textbf{Kitti\_12bit}}\\
            \cline{2-5} & \multicolumn{2}{c|}{bpp} & \multicolumn{2}{c|}{gain} \\
            \hline
            7 & \multicolumn{2}{c|}{4.92} & \multicolumn{2}{c|}{10.67\%} \\
            6 & \multicolumn{2}{c|}{4.00} & \multicolumn{2}{c|}{10.10\%} \\
            5 & \multicolumn{2}{c|}{3.09} & \multicolumn{2}{c|}{10.56\%} \\
            \hline                
        \end{tabularx}

    \end{minipage}  
    \hfill
    \begin{minipage}{0.38\textwidth}
        \centering
        \includegraphics[width=0.99\textwidth]{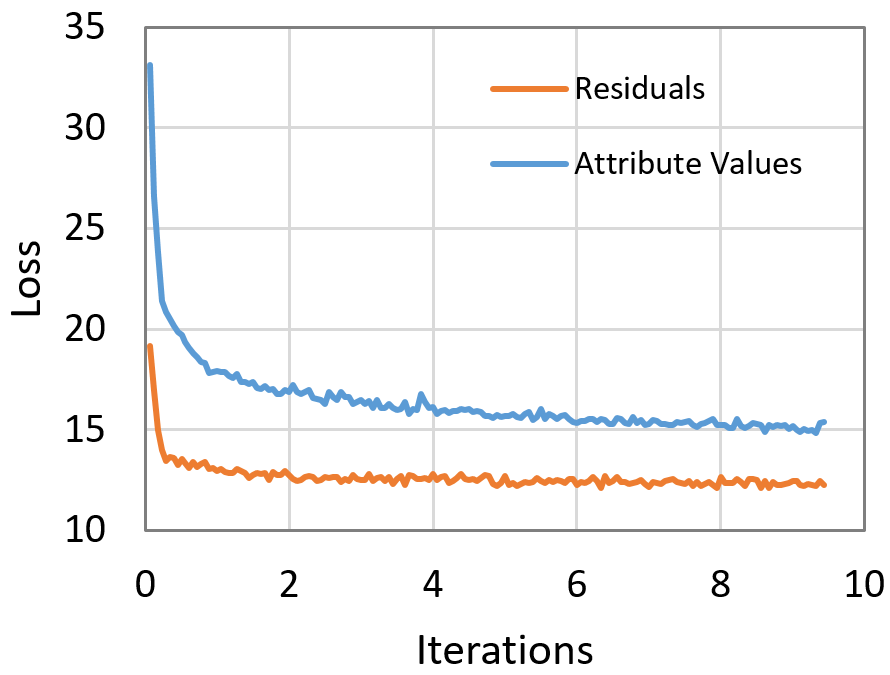}
        \caption{Convergence of attribute values vs. residuals as inputs under identical training conditions on ScanNet.}
        \label{fig:convergence}
    \end{minipage}
\end{figure}

We evaluate the time complexity of our method in comparison to G-PCC across three datasets, as summarized in Table~\ref{table:bpp1}. For the color datasets, our approach reduces both average encoding and decoding times. On the SemanticKITTI dataset, we achieve a running time comparable to that of PRAHT.
This superior performance is due to our method's parallel computing capabilities, which allow for the simultaneous processing of points within the same refinement level and the implementation of slice partitioning.

To further evaluate the scalability of the proposed HA-PCAC, we train our model using the highest precision setting and test it on point clouds of varying scales.
we first downsample the point clouds using geometric quantization step sizes of 2, 8, and 32, creating the test datasets with various density levels.  As shown in Table~\ref{table:ex2}, our method consistently outperforms PLT across these point cloud densities for both color and reflectance attributes. Notably, on the SemanticKITTI dataset, the highest performance gain reaches 16.64\% after significant downsampling. 
In addition, we assess our model on the SemanticKITTI dataset with the reflectance values quantized at different precisions. Table~\ref{table:ex2} demonstrates that our method offers 10.56\%, 10.10\%, and 10.67\% bpp reductions compared to PLT for reflectance bits of 5, 6, and 7, respectively.
This highlights the scale-invariant nature of our design, which leverages multi-resolution representations from the proposed LoD structure, along with normalization steps for both positions and attributes.


\section{Conclusion and Future work}
A lossless attribute point cloud compression method based on a novel hierarchical attention context model is proposed in this paper. By integrating the hierarchical model structure with residual learning, our method effectively achieves point cloud attribute compression. Furthermore, the proposed method enhances parallel processing capabilities and exhibits improved generalization across diverse point clouds. Extensive experimental results demonstrate that our method outperforms the state-of-the-art G-PCC method across multiple datasets. In the near future, we aim to investigate lossy attribute compression tasks within the HA-PCAC framework.

\Section{References}
\bibliographystyle{IEEEbib}
\bibliography{refs.bib}

\begin{thebibliography}{10}

\bibitem{zhang2014point}
Cha Zhang, Dinei Florencio, and Charles Loop,
\newblock ``Point cloud attribute compression with graph transform,''
\newblock in {\em IEEE International Conference on Image Processing}, 2014, pp. 2066--2070.

\bibitem{zhang2024efficient}
Jingshu Zhang, Yueru Chen, Guoqing Liu, Wei Gao, and Ge~Li,
\newblock ``Efficient point cloud attribute compression framework using attribute-guided graph {Fourier} transform,''
\newblock in {\em IEEE International Conference on Acoustics, Speech and Signal Processing}, 2024, pp. 8426--8430.

\bibitem{graziosi2020overview}
Danillo Graziosi, Ohji Nakagami, Shinroku Kuma, Alexandre Zaghetto, Teruhiko Suzuki, and Ali Tabatabai,
\newblock ``An overview of ongoing point cloud compression standardization activities: Video-based {(V-PCC)} and geometry-based ({G-PCC}),''
\newblock {\em APSIPA Transactions on Signal and Information Processing}, vol. 9, pp. e13, 2020.

\bibitem{Mammou2018Lifting}
Khaled Mammou, Alexis Tourapis, Jungsun Kim, Fabrice Robinet, Valery Valentin, and Yeping Su,
\newblock ``Lifting scheme for lossy attribute encoding in {TMC1},''
\newblock Input document m42640, ISO/IEC JTC1/SC29/WG11 (MPEG), April 2018.

\bibitem{de2016compression}
Ricardo~L De~Queiroz and Philip~A Chou,
\newblock ``Compression of {3D} point clouds using a region-adaptive hierarchical transform,''
\newblock {\em IEEE Transactions on Image Processing}, vol. 25, no. 8, pp. 3947--3956, 2016.

\bibitem{quach2022survey}
Maurice Quach, Jiahao Pang, Dong Tian, Giuseppe Valenzise, and Fr{\'e}d{\'e}ric Dufaux,
\newblock ``Survey on deep learning-based point cloud compression,''
\newblock {\em Frontiers in Signal Processing}, vol. 2, pp. 846972, 2022.

\bibitem{3cacwang2023lossless}
Jianqiang Wang, Dandan Ding, and Zhan Ma,
\newblock ``Lossless point cloud attribute compression using cross-scale, cross-group, and cross-color prediction,''
\newblock in {\em Data Compression Conference}, 2023, pp. 228--237.

\bibitem{nguyen2023lossless}
Dat~Thanh Nguyen and Andr{\'e} Kaup,
\newblock ``Lossless point cloud geometry and attribute compression using a learned conditional probability model,''
\newblock {\em IEEE Transactions on Circuits and Systems for Video Technology}, vol. 33, no. 8, pp. 4337--4348, 2023.

\bibitem{chen2020point}
Yueru Chen, Yiting Shao, Jing Wang, Ge~Li, and C-C~Jay Kuo,
\newblock ``Point cloud attribute compression via successive subspace graph transform,''
\newblock in {\em IEEE International Conference on Visual Communications and Image Processing}, 2020, pp. 66--69.

\bibitem{fu2022octattention}
Chunyang Fu, Ge~Li, Rui Song, Wei Gao, and Shan Liu,
\newblock ``{OctAttention}: Octree-based large-scale contexts model for point cloud compression,''
\newblock in {\em AAAI Conference on Artificial Intelligence}, 2022, pp. 625--633.

\bibitem{schwarz2018emerging}
Sebastian Schwarz, Marius Preda, Vittorio Baroncini, Madhukar Budagavi, Pablo Cesar, Philip~A Chou, Robert~A Cohen, Maja Krivoku{\'c}a, S{\'e}bastien Lasserre, Zhu Li, et~al.,
\newblock ``Emerging {MPEG} standards for point cloud compression,''
\newblock {\em IEEE Journal on Emerging and Selected Topics in Circuits and Systems}, vol. 9, no. 1, pp. 133--148, 2019.

\bibitem{lawder2001querying}
Jonathan~K. Lawder and Peter J.~H. King,
\newblock ``Querying multi-dimensional data indexed using the {Hilbert} space-filling curve,''
\newblock {\em ACM SIGMOD Record}, vol. 30, no. 1, pp. 19--24, 2001.

\bibitem{wu2022point}
Xiaoyang Wu, Yixing Lao, Li~Jiang, Xihui Liu, and Hengshuang Zhao,
\newblock ``Point transformer v2: Grouped vector attention and partition-based pooling,''
\newblock in {\em Advances in Neural Information Processing Systems}, 2022, pp. 33330--33342.

\bibitem{dEon2017voxelized}
Eugene d'Eon, Ben Harrison, Tim Myers, and Philip~A. Chou,
\newblock ``8i voxelized full bodies - a voxelized point cloud dataset,''
\newblock Input document WG11M40059/WG1M74006, ISO/IEC JTC1/SC29 Joint WG11/WG1 (MPEG/JPEG), July 2017.

\bibitem{dai2017scannet}
Angela Dai, Angel~X Chang, Manolis Savva, Maciej Halber, Thomas Funkhouser, and Matthias Nie{\ss}ner,
\newblock ``{ScanNet}: Richly-annotated {3D} reconstructions of indoor scenes,''
\newblock in {\em IEEE Conference on Computer Vision and Pattern Recognition}, 2017, pp. 5828--5839.

\bibitem{geiger2012we}
Andreas Geiger, Philip Lenz, and Raquel Urtasun,
\newblock ``Are we ready for autonomous driving? the {KITTI} vision benchmark suite,''
\newblock in {\em IEEE Conference on Computer Vision and Pattern Recognition}, 2012, pp. 3354--3361.

\bibitem{TMC13v23}
MPEGGroup,
\newblock ``Geometry based point cloud compression ({G-PCC}) test model version 23.0 ({TMC13v23}) [online],'' {https://github.com/MPEGGroup/mpeg-pcc-tmc13}, 2023.

\end{thebibliography}

\end{document}